\documentclass[journal]{IEEEtran}
\usepackage[table]{xcolor}
\usepackage{color}
\usepackage{CJKutf8}
\usepackage{graphicx}
\usepackage{multirow}
\usepackage{makecell}
\usepackage{hyperref}
\usepackage{epstopdf}
\usepackage{graphicx}
\usepackage[table]{xcolor}
\ifCLASSINFOpdf
\else
\fi
\begin{document}

\title{Multi-Agent Systems: From Classical Paradigms to Large Foundation Model-Enabled Futures}
% note positions of comMASs and nonbreaking spaces ( ~ ) LaTeX will not break
%~\IEEEmembership{Fellow,~OSA,
\author{Zixiang~Wang,
        Mengjia~Gong,
        Qiyu~Sun,
        Jing~Xu,~\IEEEmembership{Senior~Member,~IEEE},
        Shuai~Mao,
        Xin~Jin,
        Qing-Long~Han,~\IEEEmembership{Fellow,~IEEE},
        and~Yang~Tang,~\IEEEmembership{Fellow,~IEEE}% <-this % stops a space
\thanks{This work was supported in part by the National Natural
Science Foundation of China (62233005, U2441245, U25B6002, 62503247) and Natural Science Foundation of Jiangsu Province (BK20230605). (Z. Wang and M. Gong contributed equally to this work. \textit{Corresponding authors}: Jing Xu, Qing-Long Han, Yang Tang.)}

%\thanks{M. Shell is with the Department
%of Electrical and Computer Engineering, Georgia Institute of Technology, Atlanta,
\thanks{Z. Wang, M. Gong, J. Xu, and Y. Tang are with the Key Laboratory of Smart Manufacturing in Energy Chemical Process, Ministry of Education, East China University of Science and Technology, Shanghai 200237, China (e-mail: y20250075@mail.ecust.edu.cn; 23012887@mail.ecust.edu.cn; jingxu@ecust.edu.cn; yangtang@ecust.edu.cn).}
\thanks{Q. Sun is with the School of Information Science and Engineering, East China University of Science and Technology, Shanghai 200237, China (email: qiyu\_sun@ecust.edu.cn).}
\thanks{S. Mao is with the School of Electrical Engineering and Automation, Nantong University, Nantong 226019, China (e-mail: mshecust@163.com).}
\thanks{X. Jin is with the Research Institute of Intelligent
Complex Systems, Fudan University, Shanghai 200433, China (xinjin@fudan.edu.cn)}
\thanks{Q.-L. Han is with the School of Science, Computing and Engineering Technologies, Swinburne University of Technology, Hawthorn VIC 3122, Australia (e-mail: qhan@swin.edu.au).}}
%\thanks{Manuscript received April 19, 2005; revised September 17, 2014.}}

%\thanks{M. Shell is with the Department
%of Electrical and Computer Engineering, Georgia Institute of Technology, Atlanta,

%\thanks{Manuscript received April 19, 2005; revised September 17, 2014.}}


\markboth{IEEE/CAA JOURNAL OF AUTOMATICA SINICA,~Vol.~X, No.~X, X~X}%
{Shell \MakeLowercase{\textit{et al.}}: Bare Demo of IEEEtran.cls
for Journals}



\maketitle

\begin{abstract}
With the rapid advancement of artificial intelligence, multi-agent systems (MASs) are evolving from classical paradigms toward architectures built upon large foundation models (LFMs).
This survey provides a systematic review and comparative analysis of classical MASs (CMASs) and LFM-based MASs (LMASs).
First, within a closed-loop coordination framework, CMASs are reviewed across four fundamental dimensions: perception, communication, decision-making, and control.
Beyond this framework, LMASs integrate LFMs to lift collaboration from low-level state exchanges to semantic-level reasoning, enabling more flexible coordination and improved adaptability across diverse scenarios.
Then, a comparative analysis is conducted to contrast CMASs and LMASs across architecture, operating mechanism, adaptability, and application. 
Finally, future perspectives on MASs are presented, summarizing open challenges and potential research opportunities.
\end{abstract}


% \fi
\IEEEpeerreviewmaketitle

\section{Introduction}
\IEEEPARstart{M}{u}lti-agent systems (MASs) have become a core artificial intelligence research paradigm with broad applications in multiple disciplines, including robotics~\cite{mandi2024roco}, social intelligence~\cite{park2023generative}, and satellite systems~\cite{FAN2025155}.
Inspired by biological swarms and functional requirements of complex distributed systems~\cite{bojappa2025review,cao2012overview}, MASs focus on how multiple autonomous agents achieve global coordination or collective intelligence through interaction~\cite{chen2025confluence}.
Compared with single-agent systems, MASs provide a natural framework for modeling complex interactions and coordination among multiple autonomous entities in real-world environments~\cite{stone2000multiagent,tran2025multi}.

In this survey, MASs that do not incorporate large foundation models (LFMs) are collectively referred to as classical MASs (CMASs)~\cite{olfati2007consensus,huh2023multi}.
CMASs rely on explicitly designed system models or task-specific learning mechanisms.
From a methodological perspective, existing CMASs research can be broadly categorized into model-based and learning-based approaches.
Model-based research has gradually established several classical problem domains and theoretical frameworks, including consensus control~\cite{zhang2025distributed}, formation control~\cite{peng2026distributed}, task scheduling~\cite{xu2026cooperative}, and bio-inspired optimization~\cite{chakraborty2017swarm}.
These studies typically assume modelable systems and clear objectives to ensure provable stability and performance~\cite{olfati2007consensus}.
However, in scenarios with unmodelable environments, unknown system dynamics and partial observability, reliance on explicit modeling and control design is often limited.
As a result, learning-based methods such as multi-agent reinforcement learning (MARL) have emerged as an important alternative, enabling agents to learn coordinated policies through interaction without accurate models~\cite{li2022applications}.
While this paradigm partially mitigates model dependency in complex settings,  MARL still suffers from limitations in sample efficiency, stability, interpretability, and generalization~\cite{yuan2023survey}. 

\begin{figure*}[h]
    \centering
    \includegraphics[width=1\linewidth]{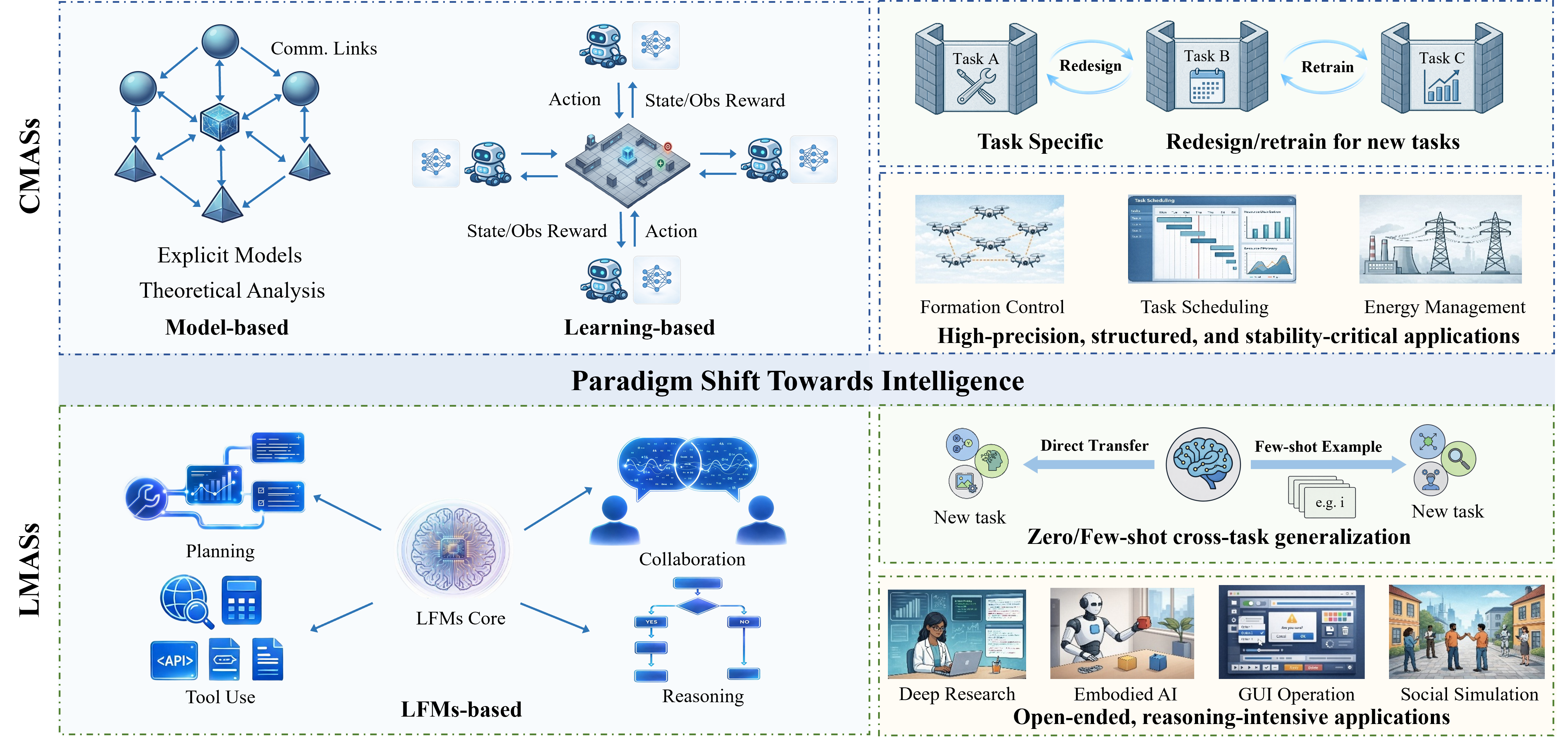}
    \caption{
    An overview of the paradigm shift from CMASs to LMASs. CMASs rely on model-based or learning-based coordination and are typically task-specific and structure-dependent. LMASs leverage LFMs to enable planning, reasoning, and zero/few-shot generalization, thereby supporting more open-ended and complex applications.
    }
    \label{fig:overview}
\end{figure*}

The limitations of CMASs motivate the exploration of more general approaches with reasoning capabilities, leading to the integration of large foundation models (LFMs) into MASs~\cite{guo2024large}.
%In recent years, LFMs have been widely integrated into MASs~\cite{guo2024large}.
In the context of MASs, LFMs serve as the cognitive core of agents.
They enable agents to interpret unstructured multimodal inputs, maintain contextual understanding, reason over complex tasks, and generate high-level actions or interaction messages~\cite{xi2025rise}.
This shifts agent operation away from predefined system models, handcrafted rules, or task-specific policies in CMASs toward semantic-level perception and language-based interaction, enabling more flexible coordination~\cite{wang2025large}.
This evolution marks a fundamental paradigm shift from task-specific, environment-constrained CMASs to more adaptive, general-purpose, and cognitively empowered LFM-based MASs (LMASs).
By leveraging the pretrained knowledge and reasoning abilities of LFMs, these systems can perform complex multi-step planning, knowledge retrieval, and high-level decision-making~\cite{ferrag2025llm,wu2025knowledge}.
As illustrated in Fig.~\ref{fig:overview}, unlike CMASs tailored to fixed environments, LMASs generalize well and accumulate experience across tasks, supporting flexible collaboration in open, dynamic scenarios~\cite{zhang2025landscape,abou2025agentic}.

The existing surveys on LMASs primarily focus on the LFM-based paradigm and summarize system architectures, coordination mechanisms, and applications~\cite{tran2025multi,ferrag2025llm,du2025survey,wang2025large,zhao2025llm}.
Unlike these surveys that examine LMASs in isolation, we propose a unified perspective bridging CMASs and LMASs.
LMASs are not a replacement for CMASs, but rather a complementary extension that enhances classical systems with high-level reasoning and generalization, while CMASs remain essential for reliable low-level control and theoretical guarantees.
Through a fine-grained, multi-dimensional analysis of their architectural and methodological connections, we aim to facilitate their synergistic co-evolution and joint advancement for future research.

The main contributions are as follows:

1) This survey provides a comprehensive overview of core theories and recent advances in MASs, covering both CMASs and LMASs.

2) CMAS and LMAS paradigms are compared from both theoretical and applied perspectives, highlighting their similarities and differences and revealing how paradigm shifts reshape MASs.

3) Key research challenges and potential future directions for the development of MASs are discussed.

Section II provides a structured review of CMASs across several key dimensions.
Section III offers a systematic analysis of LMASs from multiple core aspects.
Section IV presents a comparative analysis to elucidate the similarities, differences, and logic behind the paradigm shift from CMASs to LMASs.
Finally, Section V discusses future research directions and emerging trends in MASs.

\section{Classical MASs}

In nature, social organisms such as bird flocks, ant colonies, and bee swarms exhibit stable and efficient collective behaviors even without centralized control~\cite{bojappa2025review}.
Such phenomena have inspired research on swarm intelligence, whose core principle emphasizes the emergence of collective-level coordination and adaptability through decentralized mechanisms of self-organization~\cite{chakraborty2017swarm}.

\begin{figure*}
    \centering
    \includegraphics[width=1\linewidth]{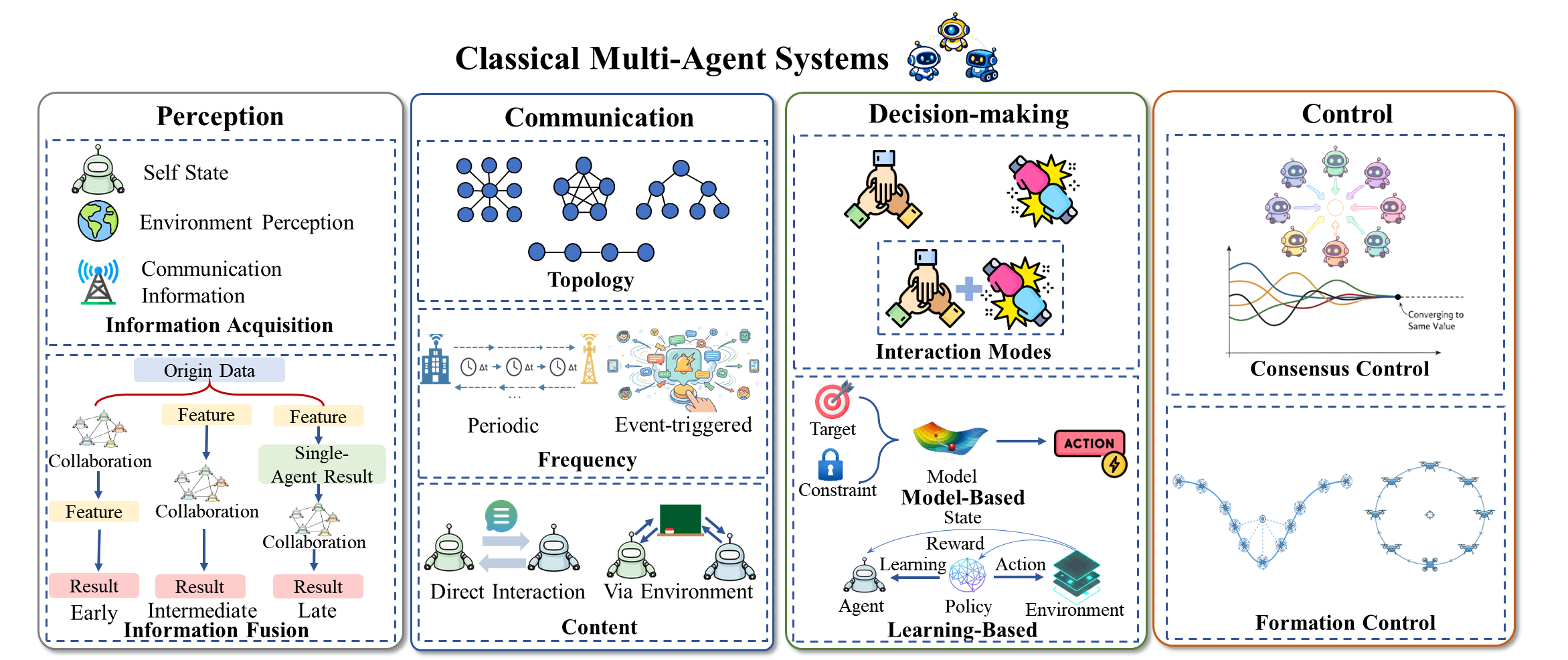}
    \caption{Classical Multi-Agent Systems: from perception and communication to decision-making, and control.}
    \label{fig:Clasic Multi-Agent System}
\end{figure*}

As shown in Fig.~\ref{fig:Clasic Multi-Agent System}, this section introduces CMASs from four aspects: perception, communication, decision-making, and control.
The first two address information acquisition and dissemination, while the latter enable distributed reasoning and coordinated actions under given objectives.
This four-dimensional framework clarifies CMASs by showing how they acquire, process, and act on information, laying the groundwork for further analysis.

\subsection{Perception}
Perception not only involves processing the agent’s own state and sensor observations, but also integrates signals transmitted by other agents~\cite{wooldridge2009introduction,chen2024computer}.
To overcome restricted views, cooperative perception relies heavily on data fusion and can be categorized into three types.
Early fusion shares raw sensor data to preserve complete perception, but demands extremely high bandwidths~\cite{chen2019cooper}.
Late fusion exchanges only abstract perception outputs after each agent completes its own perception task, making it difficult to fully exploit the potential of cooperation~\cite{yu2022dair}.
Intermediate fusion shares intermediate features, balancing communication efficiency and perception performance, and has thus become the dominant paradigm~\cite{liu2025mmcooper}.
Such fusion methods are closely related to multiscale feature aggregation widely studied in visual perception models, where shallow spatial details and deep semantic representations are integrated to improve detection performance~\cite{zhang2026novel}.
Work in this area has evolved from improving cooperative perception~\cite{yang2023spatio}, to addressing real-world perception noise~\cite{hong2024multi}, and further toward open cooperative sensing frameworks for heterogeneous agents~\cite{zhou2025pragmatic}.

\subsection{Communication}
Communication refers to the exchange of diverse information among agents to support effective collaboration for a shared goal~\cite{sun2025multi}.

From the topological perspective, communication networks are commonly represented by a graph model \(\mathbf{G} = (V, E)\), where nodes denote agents and edges represent communication links.
The network topology significantly influences the convergence and performance of the system~\cite{zhang2025network}.
The topology can either emerge from the physical deployment or be designed through heuristic or algorithmic optimization~\cite{olfati2007consensus}.
The topology can also be modeled as a learnable and dynamically adjustable process, enabling the communication structure to adaptively evolve in response to task requirements~\cite{ding2023robust}.
 
From the frequency perspective, communication can be triggered either at fixed time intervals or based on specific events~\cite{sun2025multi}.
Event-triggered communication dynamically schedules transmissions based on system states or environmental changes, thereby reducing redundant data and saving bandwidth while preserving real-time performance and control quality~\cite{hu2021event,tang2022event,jin2021event}.

From the content perspective, approaches can be divided into explicit communication through direct information exchange and implicit communication through environment-mediated interactions~\cite{ICLR2025_89b89c04}.

\subsection{Decision-making}
Decision-making governs how agents interact and coordinate, and is central to achieving system-level intelligence. 
Interaction modes can be broadly classified as cooperative, competitive, and hybrid, respectively characterized by pursuing shared goals, engaging in strategic adversarial interactions, and their combination~\cite{jin2025comprehensive}.
From a methodological perspective, decision-making approaches can further be categorized into model-based and learning-based paradigms~\cite{bojappa2025review,huh2023multi}.

Model-based decision-making relies on explicit mathematical modeling, in which the behaviors of agents are derived through methods, such as rule-based, game theory, and evolutionary optimization~\cite{bojappa2025review,panait2005cooperative,wang2025distributed}.
By modeling interactions with explicit objectives and constraints, decision strategies can be derived analytically or via optimization. 

Unlike model-based decision-making approaches, learning-based methods like MARL do not rely on precise environment models. Instead, they learn optimal or near-optimal policies through repeated interactions with the environment and the resulting reward signals~\cite{huh2023multi}.
MARL offers greater scalability and end-to-end optimization by replacing precise modeling with a learning-driven paradigm, making it well-suited for complex, partially observable, and hard-to-model environments~\cite{li2022applications}.

\subsection{Control}
In MASs, control focuses on designing distributed mechanisms to achieve coordination, stability, and global control objectives~\cite{qin2016recent}.
In this subsection, we focus on two representative control paradigms in CMASs: consensus control and formation control.

Consensus control is a fundamental problem, focusing on the design of distributed control strategies based on local information exchange to ensure that the states of all agents asymptotically converge to a common value~\cite{amirkhani2022consensus}.
Early studies primarily investigated the relationship between communication topology and consensus behavior, establishing convergence conditions under both fixed and time-varying topologies~\cite{ren2005consensus}.
Subsequently, this theory is further extended to general systems and discrete-time systems, leading to a unified analytical framework applicable to constrained communication conditions~\cite{tuna2008synchronizing,liu2011distributed}.
In recent years, learning-based methods have been adopted to learn distributed consensus control without relying on precise models, while ensuring closed-loop stability and consensus convergence~\cite{luo2023observer}.

Formation control designs control strategies that enable agents to form and maintain a predefined spatial structure during tasks~\cite{zhang2026adaptive}.
Formation control methods can be generally classified into four main categories: leader–follower, virtual structure, behavior-based, and learning-based approaches.
The leader–follower approach guides agents through designated leaders, the virtual structure approach treats the formation as a unified structure, and the behavior-based approach achieves coordination by combining primitive behaviors~\cite{oh2015survey}.
In contrast, learning-based methods enable agents to autonomously acquire formation strategies through data-driven interaction in dynamic environments~\cite{zhao2020robust}.

In this section, the four aspects are interrelated and form a closed loop.
Perception provides the informational foundation for the system.
Communication ensures the flow and sharing of information among the agents.
Decision-making coordinates actions based on this shared information.
Finally, control implements the decisions through actual actions.

\section{Large Foundation Model–Based MASs}

CMASs are mainly based on manually designed rules, explicit environment models, or learning-based training, which are often task-specific and thus limited in generalization, scalability, and adaptability~\cite{jin2025comprehensive}.
In contrast, LMASs leverage the generalization and scalability of LFMs, enabling multiple agents to collaborate on diverse requirements~\cite{wu2023brief,xiong2025deepseek,tran2025multi}.

This section reviews LMASs from four dimensions: core modules, interaction mechanisms, optimization approaches, and collective intelligence. 
These dimensions together form a unified analytical framework for understanding the structural foundations, coordination processes, and optimization strategies in open environments.

\subsection{Core Modules}
\begin{figure}
    \centering
    \includegraphics[width=1\linewidth]{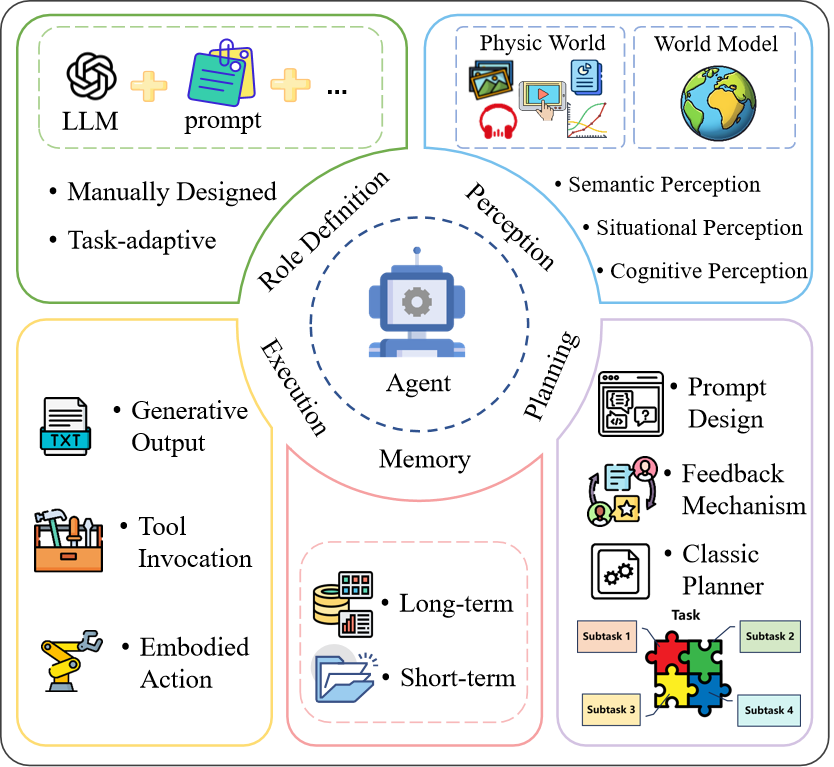}
    \caption{Core modules of an individual agent in LMASs, designed to enhance the agent’s perception, reasoning, and execution capabilities, including role definition, perception, planning, memory, execution.}
    \label{fig:core modules}
\end{figure}

As shown in Fig.~\ref{fig:core modules}, the core modules of agents in LMASs include role definition, perception, planning, memory, and execution, which provide the fundamental sources of agent capabilities.

\subsubsection{Role Definition}
Role definition assigns agents distinct responsibilities and capability boundaries to reduce redundancy and conflict, thereby facilitating complementary specialization and efficient collaboration.
Two main categories exist: manually designed roles~\cite{qian2023chatdev,ICLR2024_6507b115} and task-adaptive roles~\cite{hu2024automated,swanson2025virtual}.
Earlier LMASs rely on manually defined roles, where agent identities and functionalities are specified through predefined prompts and tool configurations.
For example, ChatDev~\cite{qian2023chatdev} and MetaGPT~\cite{ICLR2024_6507b115} decompose software development into structured subtasks and assign roles (e.g., product manager, programmer) via prompt engineering to emulate human-style division of labor.
However, manually defined roles suffer from human bias and poor generalization, motivating task-adaptive role definitions.
For example, ADAS~\cite{hu2024automated} introduces a meta-agent capable of autonomously generating, programming, and evaluating new roles, replacing static role configuration, and Virtual Lab~\cite{swanson2025virtual} focuses on dynamic team recruitment.

\subsubsection{Perception}

Prior surveys on LFM-based agents characterize perception primarily by input modalities~\cite{xi2025rise}.
We instead conceptualize agent perception as a three-level hierarchy: semantic, situational, and cognitive.
These levels represent a progression from modality-specific representation to task-aware interpretation and anticipatory understanding.
First, semantic perception maps raw sensory inputs (e.g., text, images, videos) into linguistic or symbolic representations, typically using language and vision models~\cite{devlin2019bert,7780459}. It provides the primary representations of external inputs, which are further enriched by situational and cognitive perception.
Second, situational perception conditions semantic representations on task goals, environmental states, and agent interactions to support context-aware interpretation.
For example, RoCo~\cite{mandi2024roco} integrates observations, dialogue history, and task information for coordinated planning and action.
Third, cognitive perception integrates semantic and situational information into a maintained cognitive space that supports reasoning. 
Recent work incorporates world models to extend perception with predictions of future states and environmental dynamics beyond current observations encoding~\cite{chae2024web,qiao2024agent}.

\subsubsection{Planning}
The planning module transforms high-level instructions into executable steps, enabling LFM-based agents to solve long-horizon and complex tasks~\cite{wei2025plangenllms}. 
We organize agent planning into three main aspects: structured planning generation, feedback-driven plan optimization, and planning reliability enhancement.
First, early work~\cite{wei2022chain} employs prompt engineering to impose structured reasoning, enabling models to generate intermediate steps and thus produce interpretable and scalable planning trajectories.
Second, to adapt planning trajectories to dynamic environments, feedback mechanisms are widely integrated into the planning module to iteratively optimize plans based on environmental conditions~\cite{liu2025survey}. 
Frameworks such as AutoGen~\cite{wu2023autogen} and InteRecAgent~\cite{huang2025recommender} leverage multi-agent collaboration, opinion exchange, and critic evaluation to optimize planning schemes. 
In addition, human feedback further refines planning through instructions, corrections, or preference-based guidance~\cite{liu2025survey}.
Third, to mitigate hallucination-induced errors in LFM-based planning, recent studies explore reliability enhancement strategies~\cite{wei2025plangenllms}.
These include integrating LFMs with classical task planners~\cite{zhou2024isr}, treating planning as a search process~\cite{shi2025monte}, and fine-tuning agents using task-specific planning trajectory data~\cite{hu2025agentgen}.

\subsubsection{Memory}
By analogy with human cognition, LMASs employ memory modules to store, retrieve, and update task-relevant knowledge and experience~\cite{ZHENG202524}. 
This mitigates the constraints of finite context windows by preserving valuable past experiences, thereby enhancing the quality and efficiency of subsequent reasoning and decisions~\cite{zhang2025survey}.
Memory modules consist of short-term and long-term components.
Short-term memory maintains and updates contextual information during a single task execution.
ReAct~\cite{yao2022react} interleaves reasoning traces and action histories to preserve task context.
In contrast, the long-term memory accumulates knowledge and experience across multiple executions and tasks.
Expel~\cite{zhao2024expel} implements a long-term memory mechanism that stores experiential knowledge from previous task trajectories.

\subsubsection{Execution}
Execution modules transform an agent’s internal reasoning and planning into executable actions.
LFMs achieve this through language generation, and agents can leverage external tools for improved task-specific performance~\cite{xi2025rise}.
Through the flexible combination and invocation of different tools, LFMs can achieve higher levels of autonomous execution and collaboration in complex tasks~\cite{chowa2025language}.
HuggingGPT~\cite{shen2023hugginggpt} leverages LFMs to integrate various AI models, thereby improving its capacity to solve complex tasks.
To enhance agents’ tool-use capabilities, Gorilla~\cite{patil2024gorilla} improves the tool-use performance of LFMs by constructing a large-scale API dataset and training models on it. 
With advances in embodied intelligence, agents can integrate sensory information to execute physical actions, such as navigation and manipulation, decomposing high-level instructions into actionable steps~\cite{feng2025multi,xi2025rise}.

Overall, these core modules jointly form a unified architecture that enables agents in LMASs to perform role specialization, semantic perception, structured reasoning, memory management, and flexible execution, thereby supporting scalable and adaptive multi-agent collaboration.

\subsection{Interaction Mechanisms}

\begin{figure}[h]
    \centering
    \includegraphics[width=1\linewidth]{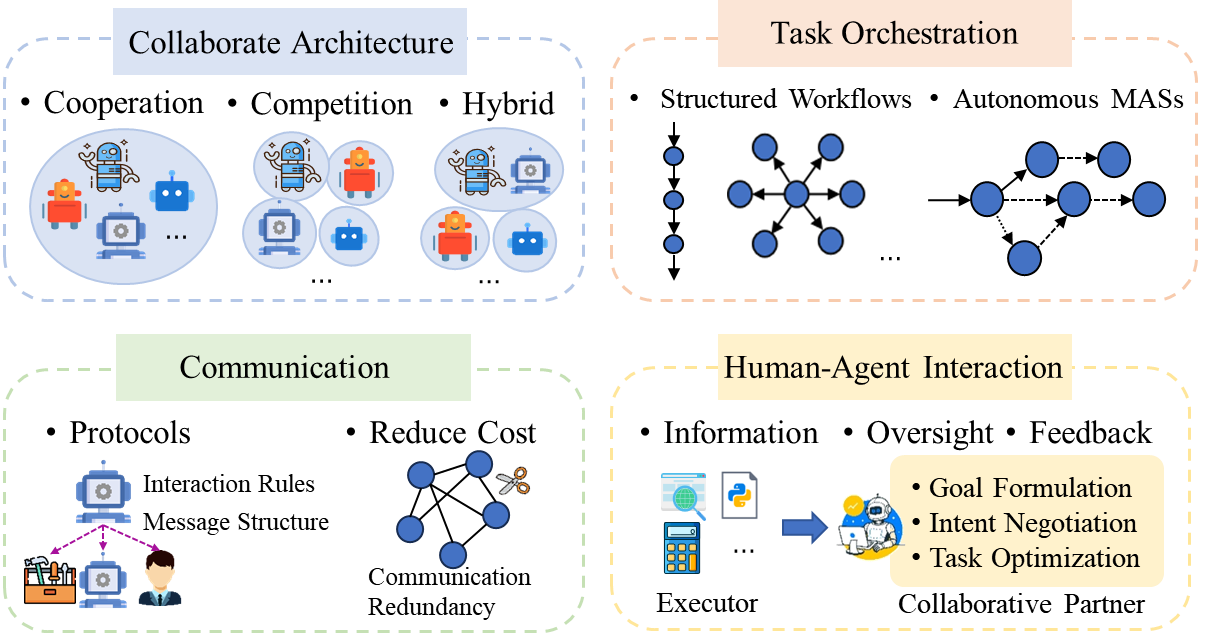}
    \caption{Interaction mechanisms in LMASs, designed to enable iterative coordination, including collaborative architecture, task orchestration, communication, and human-agent interaction.}
    \label{fig:Interaction mechanisms}
\end{figure}

The core of LMASs lies in the design of interaction mechanisms.
As shown in Fig.~\ref{fig:Interaction mechanisms}, this paper conceptualizes the interaction mechanism of MASs into four hierarchical layers: the collaborative architecture defines the interaction paradigms and organizational structures; the task orchestration manages task decomposition and execution processes; the communication mechanism supports information expression and interaction; and the human–agent interaction enables human participation and collaborative decision-making at critical stages.

\subsubsection{Collaborative Architecture}
The collaborative architecture determines how agents communicate, align their goals, and coordinate strategies at the group level~\cite{tran2025multi}. 
Based on interaction patterns, collaborative architectures can be grouped into three paradigms: cooperation, competition, and hybrid.

In the cooperative paradigm, agents align individual objectives with a shared global goal, allowing the system to decompose and assign tasks. 
Such a cooperative architecture is well-suited to workflows requiring explicit coordination and controlled execution, such as scientific experimentation~\cite{swanson2025virtual}, software programming~\cite{qian2023chatdev}.
The competitive paradigm models agent interactions under objective divergence or resource contention. 
A representative framework is multi-agent debate, where agents present opposing arguments and a judge agent selects the final result~\cite{liang2023encouraging}.
Competition can improve reasoning quality by exposing errors and alternative viewpoints.
\cite{hu2025removal} applies debate mechanisms to the retrieval and generation stages of retrieval-augmented generation (RAG), which effectively reduces hallucinations in LFMs' outputs.
The hybrid paradigm allows agents to switch between collaborative and competitive interactions based on the dynamic task stage, and is valuable in scenarios including social simulation~\cite{jinxin2023cgmi}, negotiation~\cite{zhang2025socioverse}.

\subsubsection{Task Orchestration}
Task orchestration in LMASs spans a spectrum from structured LLM workflows to more autonomous multi-agent coordination.
Many frameworks adopt structured multi-role LLM orchestration, as predefined task decomposition, role assignment, and execution protocols provide stronger controllability and stability~\cite{zhou2025reso,trirat2024automl}.
For example, ReSo~\cite{zhou2025reso} uses a directed acyclic graph–based workflow and a reward-guided selection strategy for assigning agents to subtasks.
As task openness and environmental uncertainty increase, task orchestration in LMASs is increasingly concerned with self-organization and self-evolution during execution, moving toward fully autonomous MASs.
The response-conditioned framework in~\cite{tastan2025stochastic} realizes response-driven self-organization and adaptive communication.
MAS$^2$~\cite{wang2025mas} further enables task-driven self-generation, self-configuration, and self-rectification through a meta MAS design.

\subsubsection{Communication}
Effective communication in LMASs involves two fundamental challenges: establishing structured interaction protocols and reducing communication cost. 
The first challenge concerns the design of structured communication protocols.
Communication protocols define how messages are transmitted between agents and external systems.
The Model Context Protocol (MCP)\footnote{https://modelcontextprotocol.io/introduction}, Agent-to-Agent Protocol (A2A)\footnote{https://github.com/a2aproject/A2A}, and Agent Network Protocol (ANP)\footnote{https://agent-network-protocol.com/} are among the most commonly used communication protocols. 
By standardizing message formats, semantics, and transmission procedures, these protocols organize communication structures, assign semantic roles, and optimize coordination processes, thereby enhancing the efficiency and consistency of task collaboration in MASs~\cite{yan2025beyond}. 
The second challenge focuses on minimizing communication cost.
Since interactions among agents often rely on lengthy natural language expressions, such communication can incur substantial costs in scenarios involving frequent exchanges~\cite{yan2025beyond}.
Therefore, recent studies~\cite{zhang2024cut,wang2025agentdropout} have explored pruning-based strategies to reduce multi-agent communication overhead.

\subsubsection{Human-Agent Interaction}
Human involvement in LMASs is crucial for mitigating hallucinations, goal drift, and ethical risks.
This involvement operates across information provision, supervisory governance, and iterative feedback~\cite{lu2024proactive,hua2024interactive,shao2024collaborative}.
At the information level, humans provide context, domain knowledge, and sensitive information that agents cannot reliably infer, helping to resolve ambiguities and anchor the team to unified objectives~\cite{lu2024proactive}. 
At the oversight level, humans retain ultimate decision authority to override or halt agent actions to ensure ethical and goal-consistent outcomes~\cite{hua2024interactive}.
By iterative feedback, humans guide agent teams through evaluation, correction, or implicit preference inference, shaping behavior to better align with human intent~\cite{shao2024collaborative}. 
Overall, agents support goal interpretation, refinement, and task execution under sustained human supervision~\cite{wang2025survey}. 
Importantly, they are evolving from assistive executors to collaborative partners that participate in goal formulation, intent negotiation, and task optimization~\cite{solak2025context}. 

Overall, interaction mechanisms in LMASs extend classical coordination paradigms by integrating flexible collaborative architectures, adaptive task orchestration, and semantic-based communication, while incorporating human-in-the-loop supervision. 
These mechanisms enable agents to coordinate more effectively in complex and dynamic environments, supporting a shift from structured interaction to more autonomous and intelligent collaboration.

\subsection{Hierarchical Optimization of LMASs}
\begin{figure*}
    \centering
    \includegraphics[width=1\textwidth]{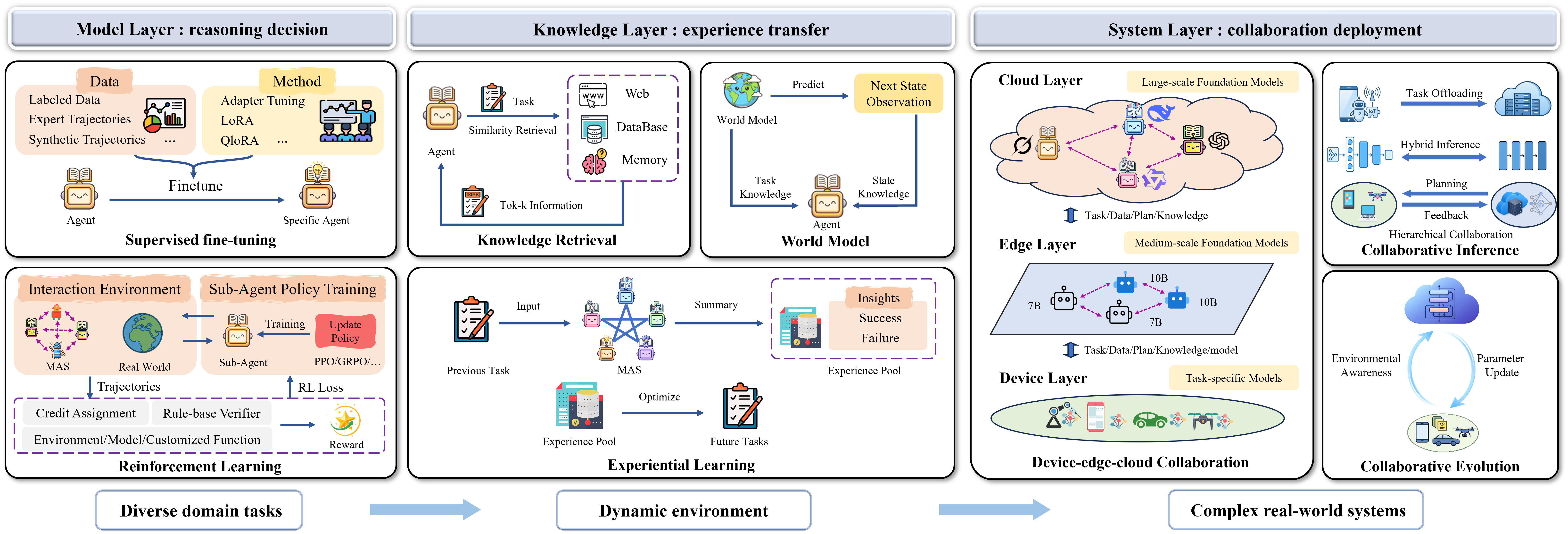}
    \caption{Hierarchical evolutionary mechanism of LMASs: through continuous optimization at the model layer, knowledge layer, and system layer, LMASs achieve adaptive evolution to accommodate diverse domain tasks, dynamic environments, and complex real-world systems. }
    \label{fig:evolution}
\end{figure*}

LMASs have shown promising results in static or well-defined settings.
However, their effectiveness is limited in dynamic environments or tasks. Therefore, optimization is crucial for enabling robust and adaptive behaviors in LMASs~\cite{du2025survey}.
As shown in Fig.~\ref{fig:evolution}, to address this, a hierarchical optimization framework spanning the model, knowledge, and system layers is needed:
(1) the model layer shapes task-specific reasoning and decision capabilities;
(2) the knowledge layer enables cross-task experience accumulation and transfer; and 
(3) the system layer enables cross-environment task execution across heterogeneous device–edge–cloud infrastructures.

\subsubsection{Model Layer}
The capabilities of LFMs directly determine LMASs' performance.
The core goal of model layer optimization is to adapt model parameters to diverse task scenarios, enabling agents to exhibit stronger robust generalization in heterogeneous tasks and dynamic environments.
This optimization is primarily achieved through supervised fine-tuning (SFT) and reinforcement learning (RL) techniques~\cite{zhang2026instruction,wang2025ragen}.

SFT is a fundamental method that optimizes pretrained models toward specific task preferences via high-quality instruction–response pairs or expert trajectories~\cite{zhang2026instruction}.
Many studies have enhanced the capabilities of intelligent agents in specific tasks through SFT.
In OS-Kairos~\cite{cheng2025kairos}, agents are fine-tuned on GUI interaction trajectories annotated with confidence, enabling step-wise confidence prediction and adaptive decisions between autonomous action and human intervention.
In AgentGen~\cite{hu2025agentgen}, agents are fine-tuned on a diverse and progressively challenging task-planning dataset to improve planning capabilities.

RL converts environmental feedback into optimization signals, enabling models to improve decision-making through multi-step interaction and long-horizon reasoning. 
In the context of single-agent RL training, long-horizon credit assignment is addressed by the StarPO framework~\cite{wang2025ragen}, where multi-turn interactions are jointly optimized.
Meanwhile, AgentGym-RL framework~\cite{xi2025agentgym} focuses on stabilizing long-horizon RL optimization by introducing a modular training framework with progressive interaction scaling.
To support joint training in multi-agent settings, Microsoft recently introduced Agent Lightning~\cite{luo2025agent}, which decouples agent execution from RL training and enables scalable, flexible optimization of LMASs.
Moreover, to overcome the reliance on predefined reward signals in existing multi-agent RL frameworks, CoMAS~\cite{xue2025comas} derives intrinsic rewards from structured inter-agent interactions to enable decentralized co-evolution.

\subsubsection{Knowledge Layer}
Knowledge-layer optimization focuses on improving the capacity of agents to acquire, accumulate, and utilize knowledge. 
It enhances multitask reasoning, collaboration, and adaptability without changing LFMs' parameters.
Current research focuses mainly on three directions: knowledge retrieval, experience learning, and world model construction. 

Knowledge retrieval is crucial because LFMs are limited by their training data and cannot directly access real-time or specialized information.
Integrating external retrieval compensates for these gaps in both timeliness and domain expertise.
External knowledge retrieval typically follows the RAG framework, which is comprehensively summarized in~\cite{gao2023retrieval} with advances spanning retrieval, augmentation, generation, and architectural design.
Recent agent systems such as TradingAgents~\cite{xiao2024tradingagents} and AutoML-Agent~\cite{trirat2024automl} incorporate external knowledge retrieval to access domain-specific and real-time information, improving task performance.

Experience learning emphasizes that agents continuously optimize and self-improve by interacting with environments, iteratively accumulating experience, and internalizing knowledge~\cite{du2025survey}. 
Reflexion~\cite{shinn2023reflexion} generates language-based self-reflection signals and stores them in memory, providing valuable self-feedback that informs future attempts. This enables the model to learn quickly and efficiently from experience.  
Expel~\cite{zhao2024expel} collects experience from a series of training tasks through trial and error, and uses successful experiences as in-context examples during the inference phase.  
Inspired by neural backpropagation, EvoMAC~\cite{hu2024self} iteratively adapts workflow structures via textual environmental feedback at test time.

A world model can be viewed as an analog of human cognitive and imaginative functions, allowing humans to internally simulate the dynamics of the external world and anticipate the potential outcomes of their actions~\cite{10522953}.
Integrating a world model into LMASs enables agents to predict future states without real-world interaction, which reduces blind exploration and hallucinations.
A parametric world model is trained in~\cite{qiao2024agent} to embed task and state knowledge into trajectories, enabling agents to leverage global task priors and local state cues for more accurate planning.
It is shown in~\cite{chae2024web} that simulating action outcomes with a world model improves web-agent decision-making.

\subsubsection{System Layer}
System-level optimization focuses on building a collaborative device–edge–cloud architecture to overcome the computational and storage challenges of deploying LFMs.
In this hierarchical setup, heterogeneous models, tasks, and resources are allocated efficiently across three layers. 
The device layer is responsible for real-time perception, preprocessing, and privacy-related tasks.
The edge layer runs medium-scale models to handle offloaded computation and coordination.
The cloud layer leverages LFMs for global planning and knowledge integration.
This paradigm is expected to enhance the collaborative inference and evolution capabilities of LMASs.

Prior works on collaborative inference across models of varying scales fall into three paradigms: task offloading, hybrid inference, and hierarchical collaboration.
Firstly, task offloading methods transfer heavy computation to cloud layers to alleviate device-side constraints~\cite{fang2023drl}.
Secondly, hybrid inference mechanisms combine large and small models to balance performance and computational cost~\cite{wang2024end}.
~\cite{fan2025madrl} enables cross-layer collaboration by partitioning models across computational layers, and~\cite{jin2025collm} introduces a confidence-aware mechanism in which low-confidence edge inference triggers cloud-based LFM assistance to preserve output quality and stability.
Finally, hierarchical architectures delegate global planning and perception to cloud models, while lightweight on-device models focus on execution~\cite{wang2025large}.
For example, EcoAgent~\cite{yi2025ecoagent} and RoboOS~\cite{tan2025roboos} both employ this design, with cloud models responsible for strategic tasks and device models handling execution and integration of multiple capabilities.

Beyond inference-time coordination, collaborative evolution is a joint learning paradigm between large cloud models and lightweight device models, aiming for coordinated optimization in real-world open environments~\cite{niu2025collaborative}.
For example, the domain adaptability of device models under dynamic environments is improved by sampling a subset of device-side data and uploading it to the cloud, where knowledge distillation and related techniques are applied, as demonstrated in~\cite{gan2023cloud,wang2024cloud}.

Overall, this hierarchical optimization framework systematically enhances LMASs across multiple levels. The model layer improves agent capabilities through model retraining. The knowledge layer supports cross-task generalization via experience accumulation and reuse. The system layer enables adaptive execution through device–edge–cloud collaboration. Together, these layers provide a unified mechanism for improving robustness and scalability in dynamic environments.

\subsection{Emergent Collective Intelligence in LMASs}
In this paper, we define collective intelligence emergence in LMASs as the appearance of qualitatively new abilities that no single agent can achieve, arising from repeated agent interactions under a shared environment and communication protocol.
At a small scale, collective intelligence primarily manifests as coordination gains that improve overall task performance.
As LMASs scale to medium- and large-scale, they give rise to population-level emergent phenomena, highlighting their potential for research in social simulation.

\subsubsection{Small-Scale Task Coordination}
Task-oriented collective intelligence arises in structured LMASs toward defined objectives, often with a relatively small number of agents, in some cases fewer than ten.
Unlike ordinary LMASs that merely distribute subtasks, small-scale emergent intelligence can give rise to behaviors such as complementary information integration and stable consensus formation.
For example, AgentVerse~\cite{chen2023agentverse} shows that modular expert teams outperform individual agents and display emergent behaviors such as spontaneous assistance and implicit goal alignment.
Some studies further examine how collective performance evolves with agent population size.
MACNET~\cite{qian2024scaling} explores collaboration from a few to thousands of agents and identifies a collaborative scaling law. System performance grows logistically with agent count, and collaborative emergence occurs earlier than traditional neural scaling.

However, LFM-based agents inherently carry risks of role inconsistency, accumulation of cognitive biases, and large-scale coordination inefficiencies, which can degrade task performance and hinder reliable collective decision-making~\cite{liu2025advances}. 
This underscores that autonomous interaction is insufficient for reliable emergence, necessitating regulation through internal attribute constraints and group-level control mechanisms.
Targeted solutions have been proposed.
CGMI~\cite{jinxin2023cgmi} enforces role consistency through a hierarchical persona model grounded in structured personality attributes.
Explicit belief state representations are introduced in~\cite{li2023theory} to reduce reasoning errors caused by hallucination and long-context degradation.

\subsubsection{Medium to Large-Scale Social Simulation}
Medium to large-scale social simulation involves tens to millions of agents interacting without predefined global objectives. 
It is driven by coordination among numerous small agent groups, and scaling these interactions to open-ended settings gives rise to population-level emergent dynamics.
The Smallville sandbox world is a canonical example~\cite{park2023generative}.
It simulates an interactive town with autonomous generative agents possessing memory and planning capabilities, giving rise to emergent social phenomena such as information diffusion, relationship formation, and spontaneous coordination.
Recent social simulators such as OASIS~\cite{yang2024oasis} scale to one million agents, revealing diverse social dynamics, including group polarization and herd behavior.
SocioVerse~\cite{zhang2025socioverse}, built upon a 10 million user pool, demonstrates predictive accuracy in simulations of presidential elections, news feedback, and economic surveys.

Overall, emergent collective intelligence highlights that the capability boundary of LMASs lies not only in individual agent performance, but more importantly in the formation and regulation of collective behaviors. This perspective shifts the focus of LMAS research from optimizing isolated agents to understanding how collective behaviors emerge, persist, and can be governed in complex environments.

\subsection{Case Study: OpenClaw}
Using OpenClaw\footnote{https://github.com/openclaw/openclaw} as an example, this section presents a case study to illustrate the practical instantiation of LMASs in real-world settings.
OpenClaw is a personal AI assistant centered on a single-agent architecture. Its modular and extensible design further enables its organization into MASs with differentiated functional roles.
This section examines OpenClaw across three levels: individual autonomy, collaborative expansion, and collective emergence.
It further discusses its behavioral characteristics under different organizational settings as well as the associated security risks and deployment implications.

At the single-agent level, OpenClaw integrates external information perception, skill usage, local operation, memory management, and autonomous optimization within a unified framework.
This integrated design enables the agent to transform users' instructions into concrete task execution.
For example, within the scope of user authorization, the agent can perform a variety of everyday digital tasks, such as automated email handling, online shopping assistance, and smart home device management.

When task complexity further increases, OpenClaw can also be organized as MASs, in which different agents assume responsibilities such as planning, retrieval, execution, and verification.
For example, in smart home device management tasks, coordinator agents assign tasks, perception agents gather  environmental information, and execution agents operate specific household devices such as robotic vacuums.
Communication agents then report results to users and receive corrective instructions.
Through this collaborative mechanism, the system can complete tasks more efficiently while demonstrating the potential to extend toward multi-agent collaboration.

At the level of collective emergence, sustained autonomous interaction among OpenClaw agents in the human-unmoderated Moltbook\footnote{https://www.moltbook.com/} community gives rise to role differentiation and functional stratification~\cite{de2026collective}.
These agent populations not only form interaction networks around shared topics, but also spontaneously construct self-organizing governance structures resembling religions and republics, reflecting clear emergent convention formation.
Although organized multi-agent collaboration trajectories can arise spontaneously, overall task performance remains fragile, and robust autonomous coordination has yet to be fully realized~\cite{yee2026molt}.

OpenClaw extends risk beyond text generation into real-world action.
As it continuously maintains and manages users’ historical records, files, account information, and operational context, it inherently becomes a concentration point for highly sensitive data.
Once affected by prompt injection attacks or unreliable tool outputs, it may produce irreversible real-world consequences, including privacy breaches, unauthorized access, and unintended deletion or modification.
Furthermore, the substantial token and computational costs of persistent online operation and iterative tool use suggest that such autonomous agents should be evaluated not only for security, but also for sustainability.

In this section, four analytical dimensions are interrelated to form a comprehensive analytical framework.
The core module provides the foundational structure for the system. 
The interaction mechanism facilitates coordination and collaboration among agents. Optimization methods enhance the system's performance in adapting to different scenarios. 
Collective intelligence demonstrates the emergent behaviors among agents.
Through this framework, the collaborative and intelligent decision-making capabilities of LMASs are gradually revealed.
In addition, the OpenClaw case study provides a practical example of the application of MASs in real-world settings.

\section{Comparative Analysis of CMASs and LMASs}

\begin{table*}[t]
\centering
\caption{Comparison of similarities and differences between CMASs and LMASs (red: similarities; blue: differences).}
\label{tab:classic_vs_llm_MASs}
\renewcommand{\arraystretch}{1.25}
\setlength{\tabcolsep}{6pt}
\begin{tabular}{p{2cm} p{2.5cm} p{5.7cm} p{5.7cm}}
\hline
\textbf{Dimension} & \textbf{Aspect} & 
\multicolumn{1}{c}{\textbf{Classical Multi-Agent Systems}} &
\multicolumn{1}{c}{\textbf{LFM-based Multi-Agent Systems}} \\
\hline

\multirow{7}{*}{\textbf{Architecture}} &
\cellcolor{red!5}\textbf{Properties} &
\multicolumn{2}{{>{\columncolor{red!5}}c}}{Autonomy; reactivity; initiative; and social ability} \\
&\cellcolor{red!5}\textbf{Objectives} &
\multicolumn{2}{>{\columncolor{red!5}}c}{Shared system-level goals}
\\ 
& \cellcolor{red!5}\textbf{Organizational} &
\multicolumn{2}{>{\columncolor{red!5}}c}{Centralized; decentralized; hybrid}  \\ 
& \cellcolor{red!5}\textbf{Execution} &
\multicolumn{2}{>{\columncolor{red!5}}c}{Parallel; distributed} \\ 
& \cellcolor{red!5}\textbf{Emergence} &
\multicolumn{2}{>{\columncolor{red!5}}c}{Beyond individual agent capabilities via interaction} \\ 

& \cellcolor{blue!5}\textbf{Individual ability} &
\cellcolor{blue!5}\makecell[c]{Low} &
\cellcolor{blue!5}\makecell[c]{High} \\

& \cellcolor{blue!5}\textbf{Deployment}  &
\cellcolor{blue!5}\makecell[c]{High-frequency real-time control; inference-light} &
\cellcolor{blue!5}\makecell[c]{Compute-intensive; lower-frequency} \\
\hline

\multirow{7}{*}{\textbf{Mechanisms}} &
\cellcolor{red!5}\textbf{Interaction} &
\multicolumn{2}{>{\columncolor{red!5}}c}{Closed-loop perception-decision-action; Information exchange} 
\\ 
&\cellcolor{red!5}\textbf{Collaboration} &
\multicolumn{2}{>{\columncolor{red!5}}c}{Cooperation; competition; hybrid} \\ 
&\cellcolor{blue!5}\textbf{Perception} &
\cellcolor{blue!5}\makecell[c]{Structured inputs} &
\cellcolor{blue!5}\makecell[c]{Unstructured multimodal inputs}  \\
& \cellcolor{blue!5}\textbf{Action} &
\cellcolor{blue!5}\makecell[c]{Low-level control; fine-grained}&
\cellcolor{blue!5}\makecell[c]{High-level intent; reasoning} \\
& \cellcolor{blue!5}\textbf{Communication} &
\cellcolor{blue!5}\makecell[c]{Compact; symbolic; and low bandwidth cost}&
\cellcolor{blue!5}\makecell[c]{Language; context-heavy; higher bandwidth cost} \\
& \cellcolor{blue!5}\textbf{Safety \& Reliability} &
\cellcolor{blue!5}\makecell[c]{Noise; communication failure; training instability} &
\cellcolor{blue!5}\makecell[c]{Hallucination; drift; misunderstanding; misuse} \\
\hline

\multirow{2}{*}{\textbf{Adaptability}} &
\cellcolor{red!5}\textbf{Robustness} &
\multicolumn{2}{>{\columncolor{red!5}}c}{Robustness against single points of failure}\\ 
& \cellcolor{blue!5}\textbf{Generalization} &
\cellcolor{blue!5}\makecell[c]{Redesign/retrain for new tasks} &
\cellcolor{blue!5}\makecell[c]{Zero-/few-shot cross-task transfer} \\
\hline

\multirow{3}{*}{\textbf{Applications}}
& \cellcolor{blue!5}\textbf{Advantages} &
\cellcolor{blue!5}\makecell[c]{High-precision, stability, verifiability} &
\cellcolor{blue!5}\makecell[c]{Flexible reasoning, tool-use, generalization}\\
& \cellcolor{blue!5}\textbf{Typical Scenarios} &
\cellcolor{blue!5}\makecell[c]{Formation control; task scheduling; \\ energy management} &
\cellcolor{blue!5}\makecell[c]{Embodied AI; GUI operation;\\ software engineering; social simulation} \\
\hline

\end{tabular}
\end{table*}

To elucidate the evolutionary trajectory of MASs from classical paradigms to LFM-driven approaches, this section presents a multi-dimensional comparative analysis between CMASs and LMASs.
We examine four key dimensions: architecture, operating mechanism, adaptability, and application.
Table~\ref{tab:classic_vs_llm_MASs} summarizes the comparison of similarities and differences between CMASs and LMASs across these dimensions.
By systematically dissecting the similarities and differences between classical approaches and emerging innovations, we aim to clarify the logic of technological iteration and development trends.

\subsection{Architecture}
For architecture, CMASs and LMASs share similarities across five dimensions.
1) Agent properties: Agents in LMASs inherit the core properties of CMASs, including autonomy, reactivity, initiative, and social ability~\cite{wooldridge2009introduction,buadicua2025contemporary}.
As a result, they act as autonomous decision-makers in interactions with the environment and other agents.
2) Objectives: Multiple autonomous agents operate as a cooperative system, interacting with the environment and other agents to achieve the shared system-level objectives.
3) Organizational structure: The architectural design flexibly supports centralized, decentralized, or hybrid architectures~\cite{wooldridge2009introduction}.
4) Task execution: Parallel or distributed strategies can be adopted when tasks are decomposable~\cite{patel2020decentralized,yan2025beyond}.
5) Emergent behavior: System-level behaviors that surpass the capabilities of any single agent emerge from local interactions among agents~\cite{chakraborty2017swarm}.

Despite these architectural similarities, significant differences exist between CMASs and LMASs.
1) Individual agent capability: CMASs are composed of agents with limited capabilities.
System intelligence depends largely on rule design and coordination mechanisms.
In LMASs, agents have strong general cognitive and reasoning abilities.
This greatly reduces the need for task-specific modeling and specialized training~\cite{zhao2025llm}.
2) Deployment paradigms: In CMASs, model-driven control methods are employed with an emphasis on high-frequency real-time closed-loop control.
Their communication and computation overhead is well controlled, which makes them well-suited for deployment in high-safety, resource-limited settings.
In learning-based MARL, inference generally incurs low computational overhead, while training requires substantial computational resources ~\cite{yuan2023survey}.
LMASs exhibit stronger cognitive coordination and decision-making capabilities, thereby reducing reliance on task-specific policy training and modeling~\cite{zhao2025llm}.
However, the local deployment of LMASs typically imposes substantial demands on computational and storage resources. 
Its reliance on LFMs significantly increases computational requirements. It also leads to higher inference latency and energy consumption, thereby constraining system responsiveness and feasibility in real-time or resource-constrained scenarios.
%However, local deployment of LMASs typically requires substantial compute and storage resources.
Under resource constraints, inference is often performed via cloud-based or hierarchical execution. This approach reduces inference frequency and results in coarser decision granularity~\cite{wang2025large}.

\subsection{Operating Mechanism}
For operating mechanisms, CMASs and LMASs share two fundamental similarities.
1) Interaction paradigm: The system operates under a closed-loop perception–decision–action process.
Agents observe the environment, make decisions based on objectives and knowledge, then act to influence the environment for new feedback~\cite{yao2022react}.
To mitigate incomplete local observations, agents exchange information and adjust their behaviors through interaction, ensuring alignment between local actions and global objectives.
2) Collaboration structure: Agents can form cooperative relationships toward shared objectives~\cite{ICLR2024_6507b115}.
They can also engage in competitive interactions over resources, rewards, or decision authority~\cite{hu2025removal}.
In multi-objective tasks, agents can exhibit hybrid interaction patterns in which cooperation and competition coexist~\cite{jinxin2023cgmi}.

We organize the differences between the two paradigms into four main aspects.
1) Perception: CMASs generally depend on low-dimensional, structured physical or numerical state variables from a single modality. They are usually constrained by fixed input formats~\cite{du2024survey}.
LMASs extend perception to unstructured multimodal inputs, such as text and images~\cite{xi2025rise}.
2) Action: CMASs primarily focus on low-level and fine-grained control. 
Actions are typically generated through controllers, optimization methods, or policy networks~\cite{jin2025comprehensive}.
LMASs generate high-level, intent-oriented actions through language-based reasoning and tool invocation, while delegating fine-grained control to external tools~\cite{tan2025roboos}.
For instance, in embodied scenarios, vision-language-action models may generate target poses or action descriptors, but the actual motion trajectories are realized by dedicated motion controllers~\cite{ge2025filic}.
3) Communication: CMASs use compact, structured information exchange, and thus typically incur lower communication burden than LMASs, while offering more controllable network latency and packet loss. In contrast, LMASs rely on natural language for interaction and coordination. This leads to the transmission of longer semantic messages and contextual histories, resulting in increased bandwidth consumption and lower communication efficiency~\cite{zhang2024cut}.
4) Safety risks: The safety risks in CMASs primarily arise from model errors, noise, and communication failures~\cite{zhang2021physical}.
Incorporating MARL further introduces distribution shifts, training instability, and reward hacking~\cite{huh2023multi}.
In LMASs, risks appear as problems inherent to language reasoning, such as hallucination, instruction drift, contextual misinterpretation, and tool misuse. The safety, verifiability, and collaboration constraints become newly critical challenges~\cite{xi2025rise,tran2025multi}.

\subsection{Adaptability}
Adaptability reflects the capability of MASs to remain robust under disruptions and to generalize across tasks and environments.
Both CMASs and LMASs demonstrate robustness.
They maintain system robustness against single-agent failures through distinct strategies.
CMASs achieve this through the design of decentralized coordination mechanisms~\cite{olfati2007consensus}.
LMASs achieve this via hierarchical ensemble designs~\cite{wang2024mixture}, blockchain-based decentralized consensus~\cite{luo2025weighted}, and dynamic adaptive topology reconfiguration~\cite{yang2025agentnet}.

The differences between the two paradigms primarily manifest in their generalization capabilities at the task and scenario levels.
In model-based CMASs, changes in system dynamics or task objectives typically require redesign and new analytical procedures.
Learning-based methods in CMASs often rely on task-specific training distributions. They tend to overfit environmental details and show degraded performance when faced with scenario or distributional shifts~\cite{sun2024llm,li2022applications}.
Large-scale pretraining injects broad world knowledge into the model. This enables rapid zero-shot or few-shot adaptation via in-context learning~\cite{wu2023brief}.
Additionally, LMASs translate task requirements into collaborative processes by role allocation and communication.
They also use memory mechanisms to accumulate experience over time, enhancing understanding and reasoning in novel scenarios~\cite{guo2024large}.

\subsection{Application}

\begin{figure}
    \centering
    \includegraphics[width=1\linewidth]{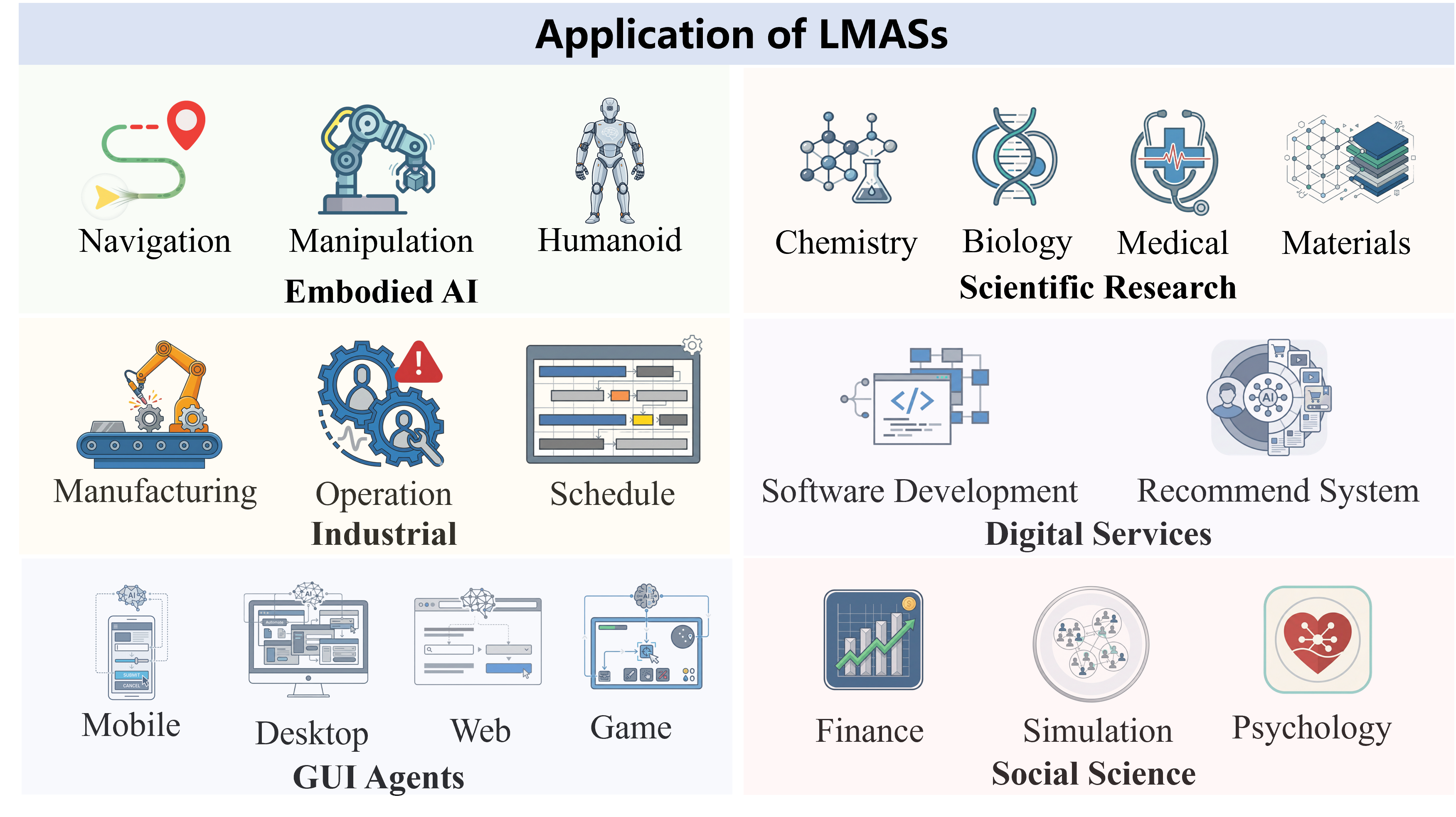}
    \caption{Broad application domains of LMASs, illustrating their versatility across virtual and physical environments.}
    \label{fig:applications}
\end{figure}

In terms of application scenarios, CMASs are well-suited for modelable, tightly constrained, and structurally defined tasks.
Model-driven approaches formulate objectives and constraints explicitly as optimization, scheduling, planning, or control problems, enabling high precision, stability, and verifiability~\cite{jin2025comprehensive}.
Learning-based methods may also be applied when the states and reward structures are well specified.
CMASs are commonly applied to tasks such as formation control~\cite{oh2015survey}, task scheduling~\cite{xu2026cooperative}, and energy management~\cite{zhao2012energy}.

In contrast, LMASs exhibit capabilities in language understanding, knowledge retrieval, and reasoning–planning, making them suitable for open-world, knowledge-intensive, and unstructured scenarios~\cite{guo2024large}.
Agents can coordinate via natural language and use tools or interfaces to ground reasoning into executable actions, enabling deployment in both virtual and physical systems. 
Compared with CMASs, LMASs demonstrate broader applicability across diverse domains.
In particular, recent agentic coding assistants such as Claude Code\footnote{https://github.com/anthropics/claude-code} and Codex\footnote{https://github.com/openai/codex} exemplify the practical deployment of LMASs in real-world applications.
They support the organization of multiple agents into collaborative teams, enabling efficient task decomposition and parallel execution.
Beyond coding assistants, LMASs have been widely explored across diverse application domains.
Fig.~\ref{fig:applications} summarizes these application scenarios across embodied robots~\cite{xu2025embodied,li2025humanoid}, industrial environments~\cite{zhang2026coma}, GUI agents~\cite{cheng2025kairos}, scientific research~\cite{ma2025medla,xiang2025artificial}, digital services~\cite{ICLR2024_6507b115}, and social science~\cite{li2025single}.

In this section, we provide a structured comparison between CMASs and LMASs to highlight their key characteristics and differences across multiple dimensions.
LMASs can be understood as both an inheritance and an extension of CMASs. 
They preserve the fundamental coordination principles of classical systems, including shared objectives, flexible organizational structures, and interaction-driven emergence. 
In addition, they extend agent capabilities from low-level control to semantic-level intent, language-based interaction, and complex task reasoning, while broadening adaptability from task-specific optimization to cross-task generalization.
Overall, LMASs represent a paradigm shift from CMASs, building upon classical foundations while extending their capabilities.

\section{Future Directions}
\begin{figure}
    \centering
    \includegraphics[width=1\linewidth]{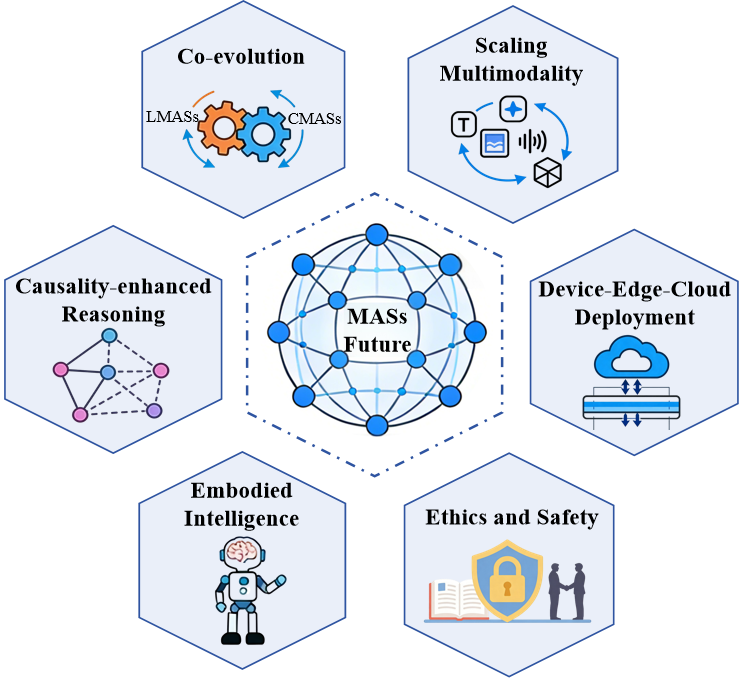}
    \caption{Future directions of generalized MASs for enhancing robustness, adaptability, interpretability, and scalability in complex open-world settings.}
    \label{fig:future direction}
\end{figure}

Future MASs are expected to operate in open, large-scale, and long-horizon environments, necessitating exploration in the following key directions (Fig.~\ref{fig:future direction}).

\subsubsection{Co-evolution of CMASs and LMASs}
CMASs and LMASs together constitute generalized MASs that integrate complementary capabilities.
Specifically, CMASs provide grounded control, stable execution, and constraint-aware coordination, while LMASs enable semantic abstraction, flexible planning, and language-mediated interaction.
The core challenge lies in aligning paradigms operating at different temporal scales and semantic levels, leading to structural mismatches between high-level reasoning and low-level control.
We therefore advocate a co-evolutionary framework of CMASs and LMASs in which classical controllers and LFM-based planners adapt jointly within shared environments, progressively aligning semantic objectives with grounded behaviors.
Effective realization of such a framework further depends on unified interfaces that connect heterogeneous agents across mismatched temporal scales and semantic grounding. These interfaces include shared perception–action abstractions, policy representations, and communication protocols.

\subsubsection{Scaling Multimodality}
If MASs rely solely on textual and visual modalities, they are susceptible to misinterpretation and suboptimal decision-making due to information loss and partial observability.
In open-world settings and real-world physical interaction tasks, it becomes necessary to incorporate modalities such as audition, tactile, and olfaction to improve perceptual robustness and reliability~\cite{xi2025rise,zhao2025polytouch}.
Future multimodal research should not only broaden the sensory modalities, but also explore a general-purpose multimodal foundation model.

Meanwhile, emerging modalities such as tactile sensing face challenges, including high costs for data acquisition and annotation, as well as complex data distributions. 
Future work could prioritize combining synthetic data generation, data augmentation, and transfer learning to build multimodal training datasets that cover multiple tasks and environments, thereby reducing reliance on expensive real-world data collection. 
In addition, modalities differ substantially in temporal scale, spatial structure, and semantic expressiveness, which can easily lead to semantic misalignment and inconsistent consensus among multiple agents. 
This calls for research on cross-modal alignment and unified representations, mapping heterogeneous inputs into a shared representation space to support stable cooperative planning and execution.

\subsubsection{Causality-Enhanced Reasoning}
In CMASs, although system states and interaction mechanisms are typically modeled explicitly, failure-propagation suppression and resource allocation often still rely on heuristic rules or empirically tuned parameters. 
In LMASs, LFM-based agents are more susceptible to being misled by superficial correlations in training data, leading to issues such as hallucinations and instruction drift. 
Meanwhile, their black-box nature makes it difficult to localize error sources, and multi-agent interactions can further amplify error propagation. 
Therefore, introducing causal models into MASs can improve the interpretability and transparency of reasoning outcomes. 
By tracing causal networks, interventions can be conducted along causal chains, enabling more robust responses to environmental changes and system perturbations.

Looking ahead, MASs could synchronously construct and update causal graphs during discussion and decision-making to constrain reasoning pathways, improving the consistency of consensus formation and the traceability of conclusions. 
Integrating interventions and counterfactual reasoning into collaborative policies would allow agents to evaluate the effects of different collaboration schemes at the action–outcome level.
This would support more robust task allocation and coordination while reducing trial-and-error costs and collaboration conflicts.
Causality-driven adaptive system reconfiguration is also a promising direction.
Agents can leverage causal relations to dynamically adjust agent topology, prompting constraints and tool configurations, so the system prioritizes key variables and decisive causal pathways and maintains stable collaborative behavior under environmental shifts or distributional drift.

\subsubsection{Device–Edge–Cloud Deployment}
Due to the limited computational and storage resources of devices, deploying MASs in real-world environments still faces significant challenges~\cite{liu2026edge}.
Devices are not only unable to host LFMs with large-scale massive parameters, but also need to satisfy strict real-time response requirements.
Therefore, constructing device–edge–cloud collaborative architectures has become an important research direction.
In this architecture, cloud-based LFMs handle high-level cognition and long-horizon task planning. 
Devices deploy lightweight models, including small-parameter foundation models and classical perception and control modules, for environmental perception and real-time execution.
In MASs, the cloud model performs task decomposition and global coordination.
On-device agents execute tasks according to their capabilities and report execution status, enabling complex long-horizon and multi-stage tasks.

Meanwhile, practical environments are often characterized by asynchronous communication and limited bandwidth.
Therefore, heterogeneous resource modeling and joint scheduling mechanisms should be developed for MASs to achieve coordinated optimization among computation, energy consumption, and latency.
In addition, robust security and privacy protection architectures should be established, along with large-scale real-world application platforms, to facilitate the scalable deployment of MASs in real environments.

\subsubsection{Embodied Intelligence}
MASs are expanding from digital spaces into the physical world through embodied intelligence~\cite{hong2026learning}. Future real-world environments will consist of large numbers of heterogeneous robots, including humanoid robots, mobile robots, robotic arms, and other specialized intelligent devices.
These robots possess complementary strengths in morphology, perception, mobility, and manipulation, enabling cooperative execution of complex tasks~\cite{liu2025coherent}.
Within this framework, each robot adopts a Brain–Cerebellum collaborative architecture~\cite{tan2025roboos}.
LFMs function as the brain, coordinating and driving embedded cerebellum-style skill modules for execution, while communication among robots further enables multi-robot collaboration.

Building on this foundation, it is necessary to further enhance the understanding and modeling of interactions among multiple embodied agents. 
For example, multi-agent world models can be constructed to characterize the interaction topology, interaction patterns, and dynamic relationships among agents and between agents and the environment, thereby improving coordination in complex scenarios.
Meanwhile, autonomous learning and self-evolution mechanisms enable agents to continually acquire reusable interaction patterns and collaboration strategies in real environments. Accumulated experience further improves collective behavior and overall system performance.
In addition, human–robot collaboration will become an important research direction, enabling embodied MASs to work with humans to accomplish complex tasks through natural interaction and safe cooperation.

\subsubsection{Ethics and Safety}
As MASs move into real-world deployments, increasing attention is being paid to whether collaborative behaviors align with human values and social norms~\cite{huang2025superalignment,gabriel2025we}. 
Compared with single-agent systems, MASs typically exhibit longer decision chains and more complex interaction patterns, which can increase ethical risks such as goal drift, fabricated or hallucinated content, and the amplification of bias. 
Future work should advance alignment from the agent level to system-level value alignment, explicitly specifying agents’ authority boundaries and information-sharing policies. 
For high-stakes tasks, stronger oversight mechanisms should be introduced, including human-in-the-loop governance, so that humans can inject critical information and exercise supervisory control at key decision points.

At the same time, ethical and governance requirements demand that MASs be auditable and accountable. 
For sensitive domains such as education, healthcare, and finance, unified compliance and responsibility frameworks are needed. 
These frameworks should translate key requirements into enforceable system rules and measurable evaluation metrics for MASs, including bias mitigation, interpretability for high-impact decisions, and accountability for errors.

\section{Conclusion}
This survey systematically examines the evolution of MASs from classical paradigms to LFM–enabled architectures. 
By analyzing CMASs and LMASs within the shared perception–communication–decision–control framework and comparing their architectures, operating mechanisms, adaptability, and applications, we clarified both their common foundations and their fundamental differences.

CMASs are grounded in explicit modeling, structured communication, and distributed control design. 
Their strengths lie in stability, verifiability, and real-time coordination under well-defined objectives and constrained environments. 
In contrast, LMASs introduce language-mediated interaction, semantic reasoning, memory augmentation, and tool integration. 
These capabilities shift coordination from low-level state synchronization toward high-level cognitive collaboration, enabling adaptation to open-ended, knowledge-intensive, and dynamically evolving tasks.
The comparative analysis reveals that the transition from CMASs to LMASs is not a replacement, but a systematic extension of coordination paradigms and intelligence capabilities.

Looking forward, future MASs research will likely focus on cross-paradigm integration, multimodality scaling, causality-enhanced reasoning, scalable device–edge–cloud deployment, embodied intelligence, and ethical issues. 
These directions collectively point toward MASs that operate across digital and physical environments, support long-horizon objectives, and maintain alignment with human oversight.

In summary, MASs are undergoing a structural transformation from rule- and model-centric coordination toward cognitively empowered, knowledge-driven collaboration.
The continued evolution of agents within MASs is expected to enable next-generation systems that are more intelligent, safer, and more efficient in complex real-world environments.

% \appendices

% % you can choose not to have a title for an appendix
% % if you want by leaving the argument blank
% \section{}

% % use section* for acknowledgment
% \section*{Acknowledgment}


\ifCLASSOPTIONcaptionsoff
  \newpage
\fi

{\small
\bibliographystyle{IEEEtran}
\bibliography{IEEEexample}
}

\end{document}